\documentclass[10pt,twocolumn,letterpaper]{article}

\usepackage{cvpr}      

\usepackage{graphicx}
\usepackage{amsmath}
\usepackage{amssymb}
\usepackage{booktabs}
\usepackage{multirow}

\usepackage[pagebackref,breaklinks,colorlinks]{hyperref}

\usepackage[capitalize]{cleveref}
\crefname{section}{Sec.}{Secs.}
\Crefname{section}{Section}{Sections}
\Crefname{table}{Table}{Tables}
\crefname{table}{Tab.}{Tabs.}

\def\cvprPaperID{4222} 
\def\confName{CVPR}
\def\confYear{2023}

\begin{document}

\title{3D-QueryIS: A Query-based Framework for 3D Instance Segmentation}

\author
{Jiaheng Liu\textsuperscript{1}, Tong He\thanks{Corresponding author.}
\textsuperscript{2}, Honghui Yang\textsuperscript{3}, 
Rui Su\textsuperscript{2}, Jiayi Tian\textsuperscript{1}, \\Junran Wu\textsuperscript{1}, Hongcheng Guo\textsuperscript{1}, Ke Xu\textsuperscript{1}, Wanli Ouyang\textsuperscript{2}\\
\textsuperscript{1}Beihang University,
\textsuperscript{2}Shanghai AI Laboratory, \textsuperscript{3}State Key Lab of CAD\&CG, Zhejiang University \\
}
\maketitle

\begin{abstract}

Previous top-performing methods for 3D instance segmentation often maintain inter-task dependencies and the tendency towards a lack of robustness. Besides, inevitable variations of different datasets make these methods become particularly sensitive to hyper-parameter values and manifest poor generalization capability.
In this paper, we address the aforementioned challenges by proposing a novel query-based method, termed as 3D-QueryIS,
which is detector-free, semantic segmentation-free, and cluster-free.
Specifically,
we propose to generate representative points in an implicit manner, and use them together with the initial queries to generate the informative instance queries.
Then, the class and binary instance mask predictions can be produced by simply applying MLP layers on top of the instance queries and the extracted point cloud embeddings.
Thus,
our 3D-QueryIS is free from the accumulated errors caused by the inter-task dependencies.
Extensive experiments on multiple benchmark datasets demonstrate the effectiveness and efficiency of our proposed 3D-QueryIS method.
\end{abstract}

\section{Introduction}
\label{sec:intro}

With the rapid development of 3D sensors and availability of large-scale 3D datasets,
3D scene understanding has received increasing attention. 
3D instance segmentation~\cite{jiang2020pointgroup,Chen_HAIS_2021_ICCV,vu2022softgroup,Schult22,he2020eccvembedding} aims to output an instance mask and its corresponding category of every object in the point clouds. It is an important perception task for 3D scene understanding, and it serves as the foundation for a wide range of applications such as robot navigation,
and autonomous driving. 

\begin{figure}[t]
\begin{center}
\includegraphics[width=\linewidth]{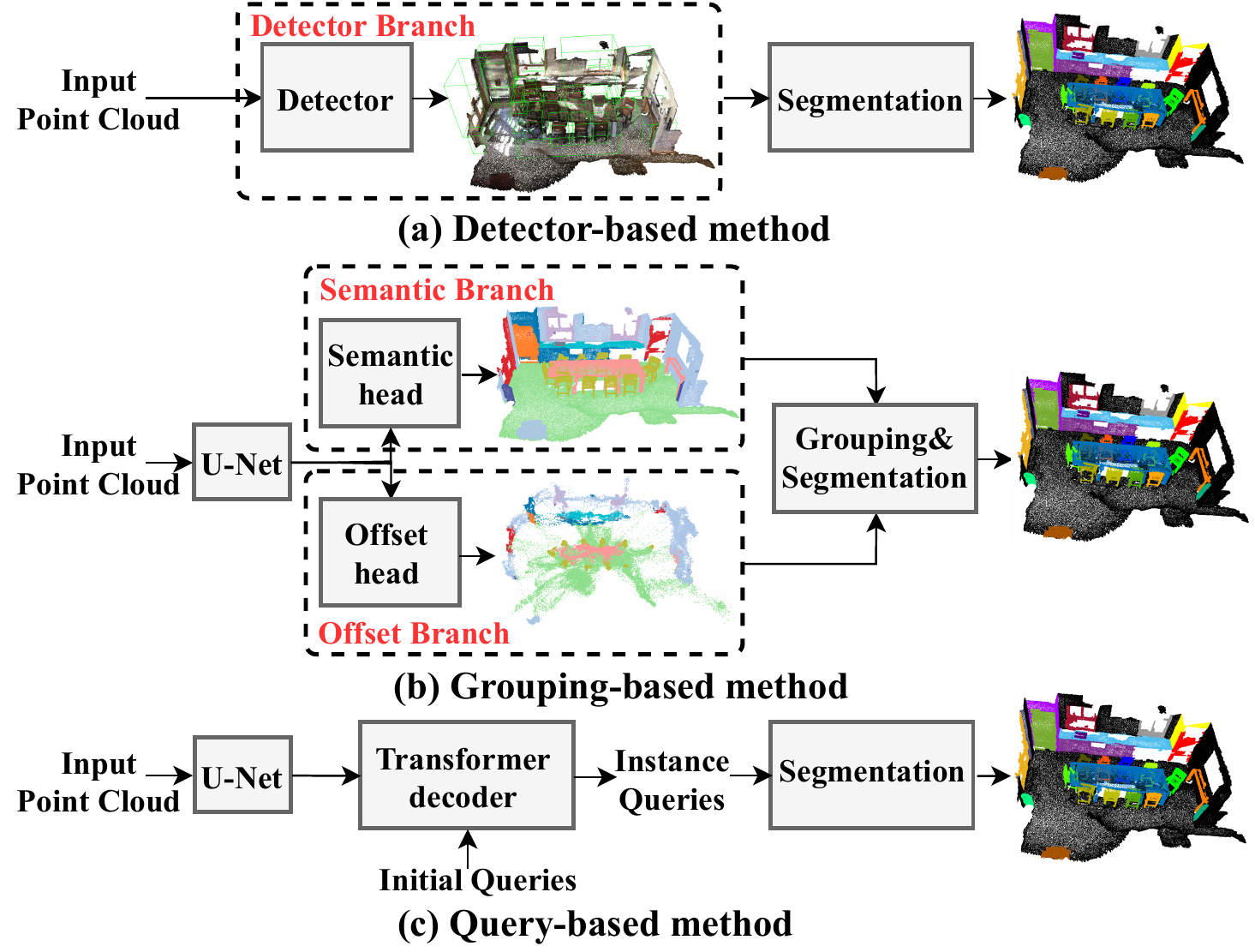}
\caption{
Different types of 3D instance segmentation methods.
 (a). The detector-based method.
(b). The grouping-based method.
(c). The query-based method (Ours).
}
\label{fig:motivatoin}
\vspace{-8mm}
\end{center}
\end{figure}

Existing 3D instance segmentation methods can be roughly divided into two categories, i.e., detector-based 3D instance segmentation methods~\cite{yi2018gspn,yang20193dbonet,yi2018gspn,hou20193dsis} and grouping-based 3D instance segmentation methods~\cite{jiang2020pointgroup,Chen_HAIS_2021_ICCV,vu2022softgroup,phamjsis3dcvpr19,liang2021instance}.
Specifically,
as shown in Fig.~\ref{fig:motivatoin}(a),
following the spirit of Mask-RCNN~\cite{he2017mask},
existing detector-based 3D instance segmentation methods~\cite{yang20193dbonet,yi2018gspn} usually adopt the top-down strategy by first using a detector to predict the bounding boxes and then segmenting the foreground points within each box.
However,
the results of these methods highly depend on the quality of the detector,
and inaccurate bounding boxes usually degrade the results a lot.
Another line of research is grouping-based methods, as shown in Fig.~\ref{fig:motivatoin}(b).
These methods first produce per-point predictions (e.g., semantic segmentation and offset prediction results) and then group points into different instances based on their features using the well-designed and heuristic cluster strategy.
However,
the performance of these methods highly depend on the quality of the clusters, which are correlated to the tasks of semantic segmentation and offset prediction.
Meanwhile, the clustering process uses many heuristic hyper-parameters,
which are sensitive and time-consuming to tune for different datasets.

Overall,
existing 3D instance segmentation methods usually introduce many inter-task dependencies (e.g., detector task or grouping task), 
which often cause sub-optimal performance due to the accumulated error. Besides, the massive heuristics make it vulnerable to unseen cases, showing limited generalization capability. 
Therefore,
it is critical to design a more elegant and dependency-less approach for 3D instance segmentation.
Motivated by the success of recent work DETR~\cite{Nicolas2020detr} on 2D detection,
we propose a new query-based framework (i.e., \textbf{3D-QueryIS}) to generate the instance queries  for 3D instance segmentation based on the transformer technique.
As illustrated in Fig.~\ref{fig:motivatoin}(c), our proposed 3D-QueryIS is detector-free, semantic segmentation-free, and cluster-free. The simplicity of our proposed 3D-QueryIS enables us to generate instance masks
without inter-task dependencies.

\begin{figure*}[t]
\begin{center}
\includegraphics[width=0.95\linewidth]{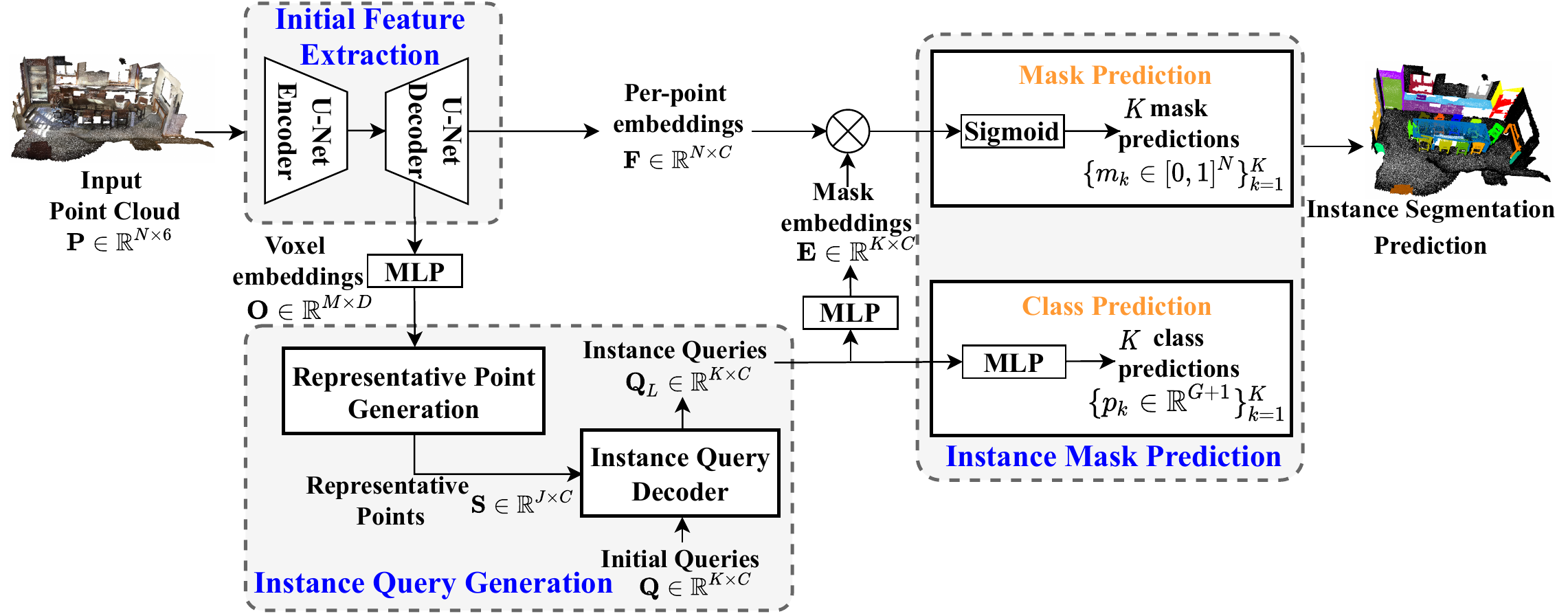}
\caption{
The overall framework of our 3D-QueryIS for 3D instance segmentation. 
3D-QueryIS includes initial feature extraction, instance query generation, and instance mask prediction. In the initial feature extraction stage,
we use U-Net to extract the per-point embeddings.
Then,
in the  instance query generation stage,
we utilize the Representative Point Generation (RPG) module to obtain the representative points,
and adopt the Instance Query Decoder (IQD) module to generate the instance queries.
Finally,
in the instance mask prediction stage,
we use the class and mask prediction heads to produce the instance segmentation results.
}
\label{fig:3dqueryinst}
\end{center}
\end{figure*}

Specifically,
our 
3D-QueryIS consists of {initial feature extraction}, {instance query generation} and {instance mask prediction} stages as shown in Fig.~\ref{fig:3dqueryinst}.
First, in the initial feature extraction stage,
we extract the per-point embedding features for each input point cloud.
Second,
in the instance query generation stage,
the Representative Point Generation (\textbf{RPG}) and Instance Query Decoder (\textbf{IQD}) modules are proposed to generate the instance queries.
For the RPG module,
 based on the voxel embeddings extracted from the U-Net decoder,
we propose to generate more informative and representative points in a learnable and implicit way,
which aims to reduce the negative effects of noise and computation costs of the transformer decoder in the IQD module.
For the IQD module,
we adopt the transformer decoder to generate the instance queries using the co-attention scheme on the generated representative points and initial queries.
Thus,
each generated instance query can  encode the global context and represent a foreground instance or background.
Third,
in the instance mask prediction stage,
based on the instance queries,
we introduce  the class prediction and mask prediction heads to generate the class and the corresponding instance mask for each instance query.

The contributions  are summarized as follows:
\begin{itemize}
    \item In our work, we investigate the problems of inter-task dependencies in existing methods,
    and propose a query-based framework called 3D-QueryIS for 3D instance segmentation, which is detector-free, semantic segmentation-free and cluster-free.

    \item 
    In the instance query generation stage,
    we propose to generate the representative points in an implicit way and produce the instance queries to encode the global context by using the transformer scheme.
    \item Extensive experiments on multiple benchmark datasets demonstrate the effectiveness of our  3D-QueryIS method. Besides, our 3D-QueryIS achieves the highest efficiency among all existing methods.
\end{itemize}
\section{Related Works}
\textbf{Point Cloud Understanding.}
Point cloud representation has been become a common data representation format for 3D scene understanding,
and a large number of deep learning methods are proposed for various point cloud understanding tasks by exploiting the geometry structure of 3D points.
For 3D recognition,
PointNet~\cite{qi2017pointnet} first utilizes a set of multi-layer perceptrons (MLPs) and a max-pooling layer to generate the global geometry representation for each point cloud.
PointNet++~\cite{qi2017pointnetplusplus} further stacks several set abstraction modules to learn the geometry representation from a set of 3D points,
which aims to exploit the local neighborhood geometry structure.
For 3D object detection,
many methods are proposed for indoor and outdoor scenes.
For example,
VoteNet~\cite{qi2019deep} adopts the backbone of PointNet++ and  the Hough voting to establish an
end-to-end 3D object detection network.
PV-RCNN~\cite{shi2020pvrcnn} utilizes both 3D voxel Convolutional Neural Network (CNN) and PointNet-based set abstraction to learn more discriminative point cloud features with manageable memory consumption.
Recently,
Transformer-based methods~\cite{vaswani2017attention} has been applied in several 3D point cloud tasks,
such as 3D object detection~\cite{misra2021end,pan20213d,chen2022focal,wu2022sparse,yang2022graphrcnn}, 3D recognition~\cite{zhao2021point,9854132},
where the Transformer model is well-suited for
operating on 3D points since they are naturally permutation invariant.

\textbf{Detector-based 3D Instance Segmentation.}
Inspired by the Mask-RCNN~\cite{he2017mask} for 2D instance segmentation,
most detector-based methods follow a top-down strategy,
which first generates the region proposals and then segments the object within each
proposal.
For example,
in GSPN~\cite{yi2018gspn}
 Li et al.~\cite{yi2018gspn} propose an analysis-by-synthesis strategy to produce high-objectness 3D proposals to handle data irregularity of point clouds.
 In 3DSIS,
Hou et al.~\cite{hou20193dsis} propose to utilize the multi-view
RGB with 3D geometry to predict both the 3D bounding boxes and instance masks.
Liu et
al.~\cite{liu2020learning} introduce the GICN to first consider the instance center of
each object as a Gaussian distribution,
and then produce the corresponding bounding
boxes and instance masks.
Yang et al.~\cite{yang20193dbonet} propose the 3D-BoNet method to directly generate the bounding boxes without using anchor generation and non-maximum suppression.

\textbf{Grouping-based 3D Instance Segmentation.} 
In contrast to the detector-based methods, 
most grouping-based methods~\cite{wang2018sgpn,phamjsis3dcvpr19,Jean2019mtml,han2020occuseg,Zhang_2021_CVPR,Chen_HAIS_2021_ICCV} adopt a bottom-up framework,
which first generates the per-point predictions (e.g., semantic maps, and geometric shifts),
and then groups the points into a set of instances.
For example,
Wang et al.~\cite{wang2018sgpn} propose the SGPN to first build a feature similarity matrix for all points and then group points
of similar features into instances. 
Pham et al.~\cite{phamjsis3dcvpr19} propose to incorporate the semantic and instance labels by a multi-value conditional random field model and jointly optimizes the labels to predict the object instances. 
To generate more accurate segmentation results,
Han et al.~\cite{han2020occuseg} introduce the OccuSeg by using the graph-based clustering guided by object occupancy signal. 
Jiang et al.~\cite{jiang2020pointgroup} propose the PointGroup to first segment objects on original and offset-shifted point sets, and then adopt a simple yet effective algorithm to group nearby points of the same label. 
Zhang et al.~\cite{Zhang_2021_CVPR} introduce
a probabilistic approach by  considering  each point as a tri-variate normal distribution followed by a clustering step to obtain object instances. 
Chen et al.~\cite{Chen_HAIS_2021_ICCV} extend PointGroup and propose HAIS to absorb surrounding fragments of instances and refine the instances based on intra-instance prediction. 

Overall, detector-based
methods process  each object proposal independently,
but the inaccurate bounding boxes degrade the instance segmentation performance a lot.
Grouping-based methods process the whole scene without using a detector, but these methods highly depend on the quality of the semantic segmentation task,
which means that the errors in semantic predictions are easily accumulated into the instance predictions.
In contrast to existing methods,
our query-based framework 3D-QueryIS aims to reduce the inter-task dependencies by using the transformer scheme,
which is  free from the influence of inaccurate bounding boxes or semantic segmentation results.

\section{Method}
In this section,
we introduce the details of our 3D-QueryIS in Fig.~\ref{fig:3dqueryinst}.
Specifically,
in the initial feature extraction stage,
for the input point cloud $\mathbf{P}$,
we follow the existing methods~\cite{vu2022softgroup,jiang2020pointgroup,Chen_HAIS_2021_ICCV,wu2022dknet} to use a U-Net style network to generate the per-point embeddings $\mathbf{F}$ and the voxel embeddings $\mathbf{O}$ generated from the middle layer of the U-Net decoder.
Then,
in the instance query generation stage,
we first generate the so-called representative points $\mathbf{S}$ using the voxel embeddings $\mathbf{O}$
and then adopt the transformer decoder to produce the instance queries $\mathbf{Q}_{L}$ based on the initial queries $\mathbf{Q}$ and the representative points $\mathbf{S}$.
Finally,
in the instance mask prediction stage,
for class prediction,
a class prediction head is used to output the class predictions using instance queries.
For the mask prediction,
we first generate the mask embeddings $\mathbf{E}$,
which are then used to take $\mathbf{F}$ as input for generating the instance masks.

\subsection{Initial Feature Extraction}
In Fig.~\ref{fig:3dqueryinst}, the input point cloud is denoted as $\mathbf{P}\in\mathbb{R}^{N\times6}$ with $N$ points,
where each point contains 3D coordinates and RGB color information.
In the initial feature extraction stage,
we first convert the unordered points
to ordered volumetric grids by voxelization, which are then fed into a U-Net style backbone implemented by sparse convolutional network~\cite{choy20194d}.
After that,
we recover points from voxels to generate the per-point embedding features $\mathbf{F}\in\mathbb{R}^{N\times C}$,
where $C$ is the feature dimension of each point.

\subsection{Instance Query Generation}
In Fig.~\ref{fig:3dqueryinst}, the instance query generation stage mainly includes two modules: Representative Point Generation (\textbf{RPG}) and Instance Query Decoder (\textbf{IQD}).
Specifically,
in RPG,
we utilize the voxel embeddings $\mathbf{O}$ from the U-Net decoder to generate the representative points $\mathbf{S}$.
In IQD,
motivated by the recent work DETR~\cite{Nicolas2020detr,cheng2021per},
we first define a set of learnable parameters
$\mathbf{Q}$ as the initial queries,
and then
exchange information together with the representative points $\mathbf{S}$ to generate the instance queries $\mathbf{Q}_{L}$
using $L$ transformer decoder layers based on the co-attention mechanism.
The details of RPG and IQD are as follows.

\noindent\textbf{Representative Point Generation}.
In order to generate instance-related queries, the co-attention mechanism is required to fuse information between the instance representations and the initial queries~\cite{Nicolas2020detr}.
{However, the whole 3D scene representations (i.e., $\mathbf{F}$) usually come with a large number of points, which can lead to a huge computational cost. 
Moreover, many points in the 3D scene are noise that might have negative effects on generating the instance-related queries.
For example, background (e.g., wall and floor) points do not correspond to any instances in ScanNet dataset~\cite{dai2017scannet}.
Meanwhile,
the authors in PointInst3D~\cite{he2022pointinst3d} observe that it is necessary to remove the background noise for generating more effective instance proposals and improving the performance of 3D instance segmentation.
As a result, sampling is needed to produce representative points from the whole scene to highlight the informative regions and reduce the computation costs.}

{One widely-used sampling approach is the task-agnostic  Farthest Point Sampling (FPS) technique,
which 
selects a group of points that are farthest apart from each other. 
However, the FPS does not take into account the further processing of the sampled points and may result in sub-optimal performance~\cite{Dovrat_2019_CVPR,Lang_2020_CVPR}.
Meanwhile,
the number of background points is usually large,
and many points sampled from FPS may also come from background regions.}
Inspired by~\cite{Cheng2022SparseInst}, we propose a representative point generation (RPG) module to generate representative points that are semantically discriminative for distinguishing instances from the 3D scene.

Specifically,
we first apply an MLP layer on top of the output from the $t$-th decoder layer in the U-Net decoder to generate the voxel embeddings 
$\mathbf{O}\in\mathbb{R}^{M\times C}$,
where $M$ is the number of voxels and $C$ is the feature dimension of each voxel.
We then apply a simple network $\mathcal{E}$ with the Sigmoid non-linearity function on $\mathbf{O}$ to generate a sparse set of instance activation maps as follows:
\begin{equation}
\label{Eq:rep1}
    \mathbf{Z} = (\mathcal{E}(\mathbf{O}))^\top,
\end{equation}
where $\mathbf{Z}\in\mathbb{R}^{J\times M}$ is a set of activation maps on the $M$ voxels, and $J$ is the number of generated activation maps.

Then,
we can obtain a sparse set of representative points $\mathbf{S}=\{s_j\}_{j=1}^{J}\in\mathbb{R}^{J\times C}$ 
by gathering distinctive information based on $\mathbf{Z}$ and the voxel embeddings $\mathbf{O}$.
For the $j$-th representative points,
the feature  $s_j$ is computed as follows:
\begin{equation}
\label{Eq:PointImportance}
    s_j = z_j \cdot\mathbf{O},
\end{equation}
where $z_j$ is the $j$-th attention maps from $\mathbf{Z}$.
After that,
the generated representative points $\mathbf{S}$ are used to generate the key and value of the  transformer decoder  for producing the instance queries in the instance query decoder module.

\noindent\textbf{Instance Query Decoder}.
We follow the existing methods~\cite{Nicolas2020detr,cheng2021per,cheng2022masked} to define a set of learnable parameters as the initial queries $\mathbf{Q}\in \mathbb{R}^{K\times C}$,
where $K$ is the number of queries and $C$ is the feature dimension.
Then,
we use the standard transformer decoder to update the queries with the co-attention mechanism using generated representative points $\mathbf{S}=\{s_j\}_{j=1}^{J}$.
The co-attention strategy of the $l$-th transformer decoder layer is illustrated as follows:

\begin{equation}
\mathbf{Q}_{l} = \mathtt{SF} \left (\frac{\mathbf{Q}_{l-1} \mathbf{K}^\top}{\sqrt{C}} \right) 
\mathbf{V}.
\label{eq:co-attention}
\end{equation}
$\mathtt{SF}$ represents the softmax operation, $H$ is the number of heads and $l\in \{1,...,L\}$,
where $L$ is the number of the transformer decoder layers.
$C$ is the number of channels in query,  key, and value.
Thus,
$\mathbf{Q}_{0}$ is the same as the initial queries $\mathbf{Q}$,
and $\mathbf{Q}_{l}$  denotes the output of the multi-head attention module in $l$-th layer. 
Meanwhile, in Eq.~\ref{eq:co-attention},
we just omit the
corresponding linear projection operations of different heads  for  query,  key, and value.
Meanwhile,
the decoder yields all predictions in parallel to preserve efficiency.

After IQD,
the generated $K$ instance queries $\mathbf{Q}_{L}\in\mathbb{R}^{K\times C}$ will encode global context,
and each instance query can be considered as one instance or background.

\subsection{Instance Mask Prediction}
In Fig.~\ref{fig:3dqueryinst},
we utilize two prediction heads (i.e., class prediction and mask prediction heads) for the instance mask prediction stage.
Specifically,
for the class prediction head,
based on the generated instance queries $\mathbf{Q}_{L}$,
we predict the class categories $\{p_k\in \mathbb{R}^{G+1}\}_{k=1}^{K}$ for $K$ queries,
where $G$ is the number of classes,
and the class prediction head generates an additional 
``no object'' category for the instance queries that do not correspond to any objects.
 For the mask prediction head, a Multi-Layer Perceptron (MLP) network with two hidden layers is used to convert the instance queries $\mathbf{Q}_{L}\in\mathbb{R}^{K\times C}$  into $K$ mask embeddings 
$\mathbf{E}=\{e_k\}_{k=1}^{K}\in \mathbb{R}^{K\times C}$,
where $e_k\in \mathbb{R}^{C}$ is the $k$-th mask embedding.
After that,
for $e_k$,
we generate the $k$-th instance mask prediction $m_k\in [0,1]^{N}$ via a dot product between 
 $e_k$ and the per-pixel embeddings $\mathbf{F}$.
The dot product is followed by a Sigmoid activation  $\sigma$ as follows:
\begin{equation}
   m_k = \sigma( \mathbf{F}\cdot e_k^\top).
\end{equation}

\subsection{Training and Inference}
\noindent\textbf{Training}.
In the training process,
 3D-QueryIS predicts a fixed-size set of predictions and it is difficult to assign ground-truth objects with hand-crafted rules.
 To tackle the end-to-end training, we formulate the label assignment using the bipartite matching~\cite{Nicolas2020detr} method,
 which has been widely used in transformer-based approaches.

Specifically,
we construct a matching cost matrix $\mathbf{C}\in\mathbb{R}^{K\times K'}$,
where $K$ is the number of queries and $K'$ is the number of ground truth instances in a scene.
For the $k$-th predicted instance query and the $k'$-th target ground-truth instance,
the matching cost $\mathbf{C}_{k, k'}$ contains the  matching cost $\mathbf{C}_{k, k'}^{ce}$ of the class prediction head and the matching cost $\mathbf{C}_{k, k'}^{mask}$ of the mask prediction head as follows:
\begin{equation}
\label{Eq:cost1}
   \mathbf{C}_{k, k'} = \lambda_{ce}\mathbf{C}_{k, k'}^{ce} + \mathbf{C}_{k, k'}^{mask},
\end{equation}
where $\lambda_{ce}$ is the cost weight of $\mathbf{C}_{k, k'}^{ce}$.
In Eq.~\ref{Eq:cost1},
 $\mathbf{C}_{k, k'}^{ce}$ is defined as $-p_k(c_{k^{'}}^{gt})$,
where $c_{k^{'}}^{gt}$ is the ground truth class of the $k'$-th target ground-truth instance.
Besides,
the matching cost $\mathbf{C}_{k, k'}^{mask}$ of the mask prediction head includes the  binary cross-entropy (BCE) (i.e., $\mathbf{C}_{k, k'}^{bce}$) and the dice matching costs (i.e., $\mathbf{C}_{k, k'}^{dice}$),
which is defined as follows:
\begin{equation}
\label{Eq:cost2}
\small
\begin{split}
   \mathbf{C}_{k, k'}^{mask} &= \lambda_{bce}\mathbf{C}_{k, k'}^{bce} + \lambda_{dice}\mathbf{C}_{k, k'}^{dice} 
   \\&= \lambda_{bce} \mathrm{BCE}(m_k, m_{k'}^{gt})+\lambda_{dice}(1-2\frac{m_k\cdot m_{k'}^{gt}+1}{|m_k|+|m_{k'}^{gt}|+1}),
   \end{split}
\end{equation}
where $\lambda_{bce}$ and $\lambda_{dice}$ are the weights of matching costs.
In Eq.~\ref{Eq:cost2},
$m_k$ and $m_{k'}^{gt}$ are the $k$-th predicted mask and the $k'$-th ground-truth mask,
respectively.

In training,
based on $\mathbf{C}_{k, k'}$,
we use Hungarian algorithm~\cite{kuhn1955hungarian} to find the optimal matching between predicted results  and
ground truth.
Then, 
 we can optimize the class and mask predictions using the following loss function:
\begin{equation}
\label{Eq:loss}
   L= \lambda_{ce} L_{ce} +\lambda_{bce} L_{mask}^{bce}+\lambda_{dice} L_{mask}^{dice}.
\end{equation}
In Eq.~\ref{Eq:loss},
we use the   cross-entropy
loss $L_{ce}$ to supervise the classification prediction,
and we consider the predicted masks that are not assigned to ground truth as the ``no object'' category.
Following~\cite{Nicolas2020detr,cheng2022masked},
the weight for the ``no object'' class in the classification loss $ L_{ce}$ is set to 0.1.
$L_{mask}^{bce}$ and $L_{mask}^{dice}$ are the binary cross-entropy loss and the dice loss between the predicted mask and the corresponding matched ground-truth mask,
respectively,
where the formulations of these losses are the same as the matching costs. 

\noindent\textbf{Inference.}
During inference,
given the input point cloud,
our 3D-QueryIS generates $K$ predicted instances with class probability predictions $\{p_k\in \mathbb{R}^{G+1}\}_{k=1}^{K}$ and corresponding instance masks $\{m_k\}_{k=1}^{K}$.
Then,
for the $k$-th instance,
following~\cite{cheng2021per},
$c_k$ is the probability of the most likely semantic class label based on $p_k$,
and the binary instance mask $b_k=m_k>\tau$,
where $\tau=0.5$ is the similarity threshold to generate the binary mask.
Finally,
the corresponding confidence score $s_k$ for the $k$-th instance is defined as follows:
\begin{equation}
\label{Eq:score}
s_k = c_k\cdot(\sum (m_k\cdot b_k))/ \sum b_k. 
\end{equation}
Intuitively, $(\sum (m_k\cdot b_k))/ \sum b_k$ represents  the confidence score for predicting the instance mask $m_k$.
When $m_k$ is more deterministic,  $(\sum (m_k\cdot b_k))/ \sum b_k$ is higher.
Thus, the confidence score $s_k$ is high when both the most likely class probability $c_k$ and the instance mask prediction probability $m_k$ are high~\cite{cheng2022masked,cheng2021per}. 
Finally,
based on $\{s_k\}_{k=1}^{K}$, $\{c_k\}_{k=1}^{K}$ and $\{m_k\}_{k=1}^{K}$,
we can generate the 3D instance segmentation results for the input point cloud. Meanwhile, Non-Maximum
Suppression (NMS) is not needed at inference in our proposed 3D-QueryIS method.

\section{Experiments}
We conduct comprehensive experiments on two standard benchmark datasets (i.e., ScanNetV2~\cite{dai2017scannet} and Stanford 3D Indoor Semantic Dataset (S3DIS)~\cite{armeni20163d}) to demonstrate
the effectiveness of our proposed 3D-QueryIS method.
\begin{table*}[t]
		\caption{Results of different methods on the validation set of the ScanNetV2 dataset. To make a fair comparison, we report the performance with different model scalability. The performance results of HAIS-S and SoftGroup-S are obtained by using their official training codes.}
	\resizebox{\textwidth}{!}{
		\begin{tabular}{l|cc|cccccccccccccccccc}
			\toprule
			&{$\mathbf{\mathrm{AP}_{50}}$} &{$\mathrm{AP}$}& \rotatebox{90}{cabinet} & \rotatebox{90}{bed} & \rotatebox{90}{chair} & \rotatebox{90}{sofa} & \rotatebox{90}{table} & \rotatebox{90}{door} & \rotatebox{90}{window} & \rotatebox{90}{bookshe.} & \rotatebox{90}{picture} & \rotatebox{90}{counter} & \rotatebox{90}{desk} & \rotatebox{90}{curtain} & \rotatebox{90}{fridge} & \rotatebox{90}{s.curtain} & \rotatebox{90}{toilet} & \rotatebox{90}{sink} & \rotatebox{90}{bath} & \rotatebox{90}{otherfu.} \\
			\midrule

			SGPN~\cite{wang2018sgpn}&11.3 &- &10.1&16.4&20.2&20.7&14.7&11.1&11.1&0.0&0.0&10.0&10.3&12.8&0.0&0.0&48.7&16.5&0.0&0.0\\
			3D-SIS~\cite{hou20193dsis}&18.7&- &19.7&37.7&40.5&31.9&15.9&18.1&0.0&11.0&0.0&0.0&10.5&11.1&18.5&24.0&45.8&15.8&23.5&12.9\\
			3D-MPA~\cite{Engelmann20CVPR}&{59.1} &35.3 &{51.9}&{72.2}&{83.8}&{66.8}&{63.0}&{43.0}&{44.5}&{58.4}&{38.8}&{31.1}&{43.2}&
			{47.7}&{61.4}&{80.6}&{99.2}&{50.6}&{87.1}&{40.3}\\
			PointGroup-S~\cite{jiang2020pointgroup} &56.9 &34.8 &48.1 &69.6 &87.7 &71.5 &62.9 &42.0 &46.2 &54.9 & 37.7 &22.4 &41.6 & 44.9&37.2 &64.4 &98.3 &61.1 &80.5 &53.0 \\
			
			DyCo3D-S~\cite{He2021dyco3d}&57.6 &35.4 &50.6 &{73.8} &84.4 & 72.1 &{69.9} &40.8 &44.5 &{62.4} & 34.8 &21.2 & 42.2 &37.0 & 41.6 & 62.7 &92.9 &61.6 & 82.6 &47.5 \\

			HAIS-S~\cite{Chen_HAIS_2021_ICCV}  &59.1 &38.0 &54.4 &76.0 &87.7 &69.4 & 66.5 & 47.5 & 48.5 & 53.1 & 43.6 &24.0 & 50.9 & {55.8} & 45.1 & 58.5 &94.7 & 53.6 & 80.8 & 53.0 \\
			SoftGroup-S~\cite{vu2022softgroup}&62.0&39.4&55.5&70.6&82.6&72.2&74.9&50.6&48.1&56.0&53.2&28.2&44.9&52.9&60.5&67.4&99.9&56.5&85.3&56.9\\
			{PointInst3D-S~\cite{he2022pointinst3d}} & 59.2&39.6  &51.1 & 75.9 & 86.5 & {72.8} & 67.3 & 45.2 & {52.3} & 57.2 & 43.8 & 25.7 & 40.5 & 53.7 & 37.2 & 59.4 & 98.2 & 58.9 & 87.0 &52.9 \\
      \textbf{3D-QueryIS-S}&\textbf{63.3}&\textbf{41.5}&50.1&74.9&87.0&65.8&76.2&55.7&50.4&52.0&47.5&58.1&59.7&50.9&47.2&63.8&98.1&60.9&83.3&57.5\\
			\midrule
			PointGroup-L~\cite{jiang2020pointgroup}&57.1&37.5&55.3&69.5&86.3&58.6&66.3&44.3&47.9&44.6&40.5&27.2&42.2&45.2&41.0&56.8&94.4&67.5&83.1&57.6\\
			DyCo3D-L~\cite{He2021dyco3d} & {61.0} & {40.6} &{52.3} & 70.4 &{90.2} & 65.8 &69.6 & 40.5 & {47.2} &48.4 &{44.7} & {34.9} & {52.3} &47.5 &51.5 & 70.3 & 94.8 & {74.3} & 77.4 &{56.4} \\
			
			HAIS-L~\cite{Chen_HAIS_2021_ICCV} & {64.0} & 43.5  &55.4 & 70.2 &82.5 &67.7 & 75.3 & 48.1 & 51.5 & 49.4 & {48.7} & {47.8} & {58.5} & 55.7 & {53.0} &76.1 & {100.0} & 69.2 & {87.1} & 56.3 \\ 
			SoftGroup-L~\cite{vu2022softgroup}&\textbf{67.6}&46.0&61.7&71.8&85.6&67.5&77.6&53.1&56.3&64.3&53.9&38.6&57.5&55.6&76.7&75.6&98.3&70.9&86.7&60.3\\
			{PointInst3D-L~\cite{he2022pointinst3d}}  &63.7 & {45.6} &{58.5} &{78.5}  &{93.6} &63.2 &{76.5}  &{55.6}  &48.5  &59.4  &38.3  &36.9  &54.2  &50.7  &46.2  &72.3 &98.3  &68.8  &{87.1}  &{59.5}  \\ 

   \textbf{3D-QueryIS-L}&66.8&\textbf{46.2}&56.8&74.8&88.4&67.7&75.4&62.2&58.4&51.7&55.7&51.9&65.6&56.1&61.9&68.5&98.5&60.6&87.6&61.7\\
			\bottomrule
		\end{tabular}
	}
	\label{tab:scannet_val}
\end{table*}
\noindent\textbf{Datasets.}
The ScanNetV2 dataset consists of 1613 scans in total, which are divided into training, validation,
and testing sets with a size of 1201, 312, and 100, respectively.
The task of 3D instance segmentation is evaluated on 18 classes.
Following existing works~\cite{vu2022softgroup,jiang2020pointgroup}, we report the results on the validation set and submit the results on the testing set to the official evaluation server to demonstrate the effectiveness of our method.
The results are reported on AP and $\mathbf{\mathrm{AP}_{50}}$ metrics.
The S3DIS dataset includes 3D scans of 6 areas with 271 scenes in total,
which is evaluated on 13 classes for 3D instance segmentation.
Following existing methods,
we report the results on two evaluation settings
(i.e., testing on Area 5 and 6-fold cross-validation).

\noindent\textbf{Implementation Details.}
We follow ~\cite{jiang2020pointgroup,vu2022softgroup} to  utilize a symmetrical U-Net as the backbone model using the sparse convolution module in the initial feature extraction stage.
The U-Net has 7 blocks in total and the channels of the block control the scalability of the model.
To show the generalization capability of our method,
we report the performance results using both small and large backbones,
which are denoted as 3D-QueryIS-S and 3D-QueryIS-L, respectively. 
Specifically,
the channel units (i.e., $C$) of U-Net for 3D-QueryIS-S and 3D-QueryIS-L are set as 16 and 32,
respectively.
The model is implemented using PyTorch framework~\cite{paszke2019pytorch} and
trained on 150k iterations with AdamW optimizer~\cite{loshchilov2018decoupled}. 
The batch size is 10,
and the learning rate is 1e-3 with a
polynomial decay policy.
In training,
$\lambda_{ce}$ in Eq.~\ref{Eq:cost1}, $\lambda_{bce}$ and $\lambda_{dice}$ in Eq.~\ref{Eq:cost2} and Eq.~\ref{Eq:loss} are set as 2, 5 and 5, respectively.
For representative point generation,
$t$  is set as 5.
and the number of representative points (i.e., $J$) is set as 2000.
For the instance query decoder,
the number of transformer decoder layers (i.e., $L$) is set as 3,
the number of heads (i.e., $H$) of the transformer decoder layer is 4,
and the number of initial queries (i.e., $K$) is set as 100.
For the class prediction,
the number of classes (i.e., $G$) is set as 18 and 13 for ScanNetV2 and S3DIS,
respectively.


\subsection{Results on ScanNet}

The performance results of 3D instance segmentation on the validation and testing sets of ScanNetV2~\cite{dai2017scannet} are reported in Table~\ref{tab:scannet_val}
and Table~\ref{tab:scannet_test}, respectively. 
{On the validation set, we compare the 3D instance segmentation performance of different methods by using both small and large backbones as defined in~\cite{he2022pointinst3d}, and our methods are termed 3D-QueryIS-S and 3D-QueryIS-L, respectively. In Table~\ref{tab:scannet_val}, when using a large backbone, our 3D-QueryIS-L can achieve a comparable results with the existing state-of-the-art method SoftGroup~\cite{vu2022softgroup}, which is a grouping-based method. However, the clustering approaches used in existing grouping-based methods (e.g., HAIS~\cite{Chen_HAIS_2021_ICCV}, SoftGroup~\cite{vu2022softgroup}) usually include many inter-task dependencies (e.g., semantic segmentation and offset prediction), and the qualities of semantic segmentation and offset prediction degrade a lot when using a small backbone. As a result, unsatisfactory grouping results are obtained and the 3D instance segmentation performance decreases. In Contrast, our 3D-QueryIS does not contain any types of dependencies, and as shown in Table~\ref{tab:scannet_val}, our 3D-QueryIS-L outperforms all state-of-the-art methods when using a small backbone. This demonstrates the benefits of removing the inter-task dependencies and the effectiveness of our framework.} 
Moreover,
we also report the results of our method on the test set in Table~\ref{tab:scannet_test}.
{It is observed that our method achieves promising results among the compared methods, which further demonstrates the effectiveness of our 3D-QueryIS.}

\begin{table*}[!t]
	
		\caption{Results of different methods on the testing set of the ScanNetV2 dataset.}
		\centering
		\resizebox{\textwidth}{!}{
		\begin{tabular}{l|c|cccccccccccccccccc}
			\toprule
			&{$\mathbf{\mathrm{AP}_{50}}$}& \rotatebox{90}{bathtub} & \rotatebox{90}{bed} & \rotatebox{90}{bookshe.} & \rotatebox{90}{cabinet} & \rotatebox{90}{chair} & \rotatebox{90}{counter} & \rotatebox{90}{curtain} & \rotatebox{90}{desk} & \rotatebox{90}{door} & \rotatebox{90}{otherfu.} & \rotatebox{90}{picture} & \rotatebox{90}{refrige.} & \rotatebox{90}{s.curtain} & \rotatebox{90}{sink} & \rotatebox{90}{sofa} & \rotatebox{90}{table} & \rotatebox{90}{toilet} & \rotatebox{90}{window} \\
			\midrule
			SGPN~\cite{wang2018sgpn}&14.3&20.8&39.0&16.9&6.5&27.5&2.9&6.9&0.0&8.7&4.3&1.4&2.7&0.0&11.2&35.1&16.8&43.8&13.8\\
GSPN~\cite{yi2018gspn}&30.6&50.0&40.5&31.1&34.8&58.9&5.4&6.8&12.6&28.3&29.0&2.8&21.9&21.4&33.1&39.6&27.5&82.1&24.5\\
3D-SIS~\cite{hou20193dsis}&38.2&100.0&43.2&24.5&19.0&57.7&1.3&26.3&3.3&32.0&24.0&7.5&42.2&85.7&11.7&69.9&27.1&88.3&23.5\\
MASC~\cite{Liu2019masc}&44.7&52.8&55.5&38.1&38.2&63.3&0.2&50.9&26.0&36.1&43.2&32.7&45.1&57.1&36.7&63.9&38.6&98.0&27.6\\
PanopticFusion~\cite{Narita2019iros}&47.8&66.7&71.2&59.5&25.9&55.0&0.0&61.3&17.5&25.0&43.4&43.7&41.1&85.7&48.5&59.1&26.7&94.4&35.9\\
3D-BoNet~\cite{yang20193dbonet}&48.8&100.0&67.2&59.0&30.1&48.4&9.8&62.0&30.6&34.1&25.9&12.5&43.4&79.6&40.2&49.9&51.3&90.9&43.9\\
MTML~\cite{Jean2019mtml}&54.9&100.0&80.7&58.8&32.7&64.7&0.4&81.5&18.0&41.8&36.4&18.2&44.5&100.0&44.2&68.8&57.1&100.0&39.6\\
3D-MPA~\cite{Engelmann20CVPR}&61.1&100.0&83.3&76.5&52.6&75.6&13.6&58.8&47.0&43.8&43.2&35.8&65.0&85.7&42.9&76.5&55.7&100.0&43.0\\
DyCo3D~\cite{He2021dyco3d}&64.1&100.0&84.1&89.3&53.1&80.2&11.5&58.8&44.8&43.8&53.7&43.0&55.0&85.7&53.4&76.4&65.7&98.7&56.8\\
PointGroup~\cite{jiang2020pointgroup}&63.6&100.0&76.5&62.4&50.5&79.7&11.6&69.6&38.4&44.1&55.9&47.6&59.6&100.0&66.6&75.6&55.6&99.7&51.3\\
OccuSeg~\cite{han2020occuseg}&67.2&100.0&75.8&68.2&57.6&84.2&47.7&50.4&52.4&56.7&58.5&45.1&55.7&100.0&75.1&79.7&56.3&100.0&46.7\\
SSTNet~\cite{liang2021instance}&69.8&100.0&69.7&88.8&55.6&80.3&38.7&62.6&41.7&55.6&58.5&70.2&60.0&100.0&82.4&72.0&69.2&100.0&50.9\\
HAIS~\cite{Chen_HAIS_2021_ICCV}&69.9&100.0&84.9&82.0&67.5&80.8&27.9&75.7&46.5&51.7&59.6&55.9&60.0&100.0&65.4&76.7&67.6&99.4&56.0\\
PointInst3D~\cite{he2022pointinst3d}&64.6&100.0&75.9&63.6&60.9&83.8&30.9&56.2&38.4&53.7&49.7&45.9&59.3&100.0&55.8&72.7&59.7&100.0&57.2\\
  
            \textbf{3D-QueryIS}&\textbf{70.5}&100.0&79.0&63.1&68.5&88.0&31.2&58.1&73.5&57.8&63.3&47.6&76.0&100.0&66.7&81.7&76.4&84.7&52.3\\
            
			\bottomrule
			
		\end{tabular}
	}
	\label{tab:scannet_test}
\end{table*}

			
			
   
	

\subsection{Results on S3DIS}
As shown in Table~\ref{tab:s3dis_ins_results},
we also provide the results of our 3D-QueryIS method on the  S3DIS dataset,
where the experiments are carried out on both Area 5 and 6-fold cross-validation following the widely-used setting~\cite{jiang2020pointgroup,Chen_HAIS_2021_ICCV}.
When compared with other existing methods,
we observe that our proposed method still achieves promising performance results on S3DIS,
which further demonstrates the effectiveness of our method.
Specifically,
when compared with the current state-of-the-art method (i.e., SoftGroup) on the testing split of Area 5,
our proposed 3D-QueryIS method obtains 3.9\%
improvements on the AP metric.

\begin{table}[!t]
	\caption{Results of different methods on the S3DIS dataset for both Area-5 and 6-fold cross validation settings.
	}
 \small
	\begin{center}
		\begin{tabular}{c|c|c|c|c}
			\toprule
			Methods  & $\mathrm{AP}$    & $\mathrm{AP}_{50}$   &  $\mathrm{mPrec}_{50}$ & $\mathrm{mRec}_{50}$ \\
			
			\midrule
			\multicolumn{5}{c}{Test on Area 5} \\
			\midrule
			SGPN ~\cite{wang2018sgpn}   &  -& -  &  36.0  & 28.7   \\
			
			ASIS~\cite{wang2019asis}   & -&  - &  55.3 &  42.4 \\
			3D-BoNet~\cite{yang20193dbonet}   & -  &  - &  57.5 &  40.2 \\
			MPNet~\cite{he2020eccvmemory}      & - &- &{62.5} &{49.0} \\
			
			PointGroup~\cite{jiang2020pointgroup}  &- &57.8& 61.9 & 62.1 \\
			DyCo3D~\cite{He2021dyco3d}  &- &- &{64.3} &{64.2} \\
			
			HAIS~\cite{Chen_HAIS_2021_ICCV} & - &- &71.1 &65.0\\
   MaskGroup~\cite{9859996}&-&65.0&62.9&64.7\\
   SSTNet~\cite{liang2021instance}&42.7&59.3&65.5&64.2\\
   SoftGroup~\cite{vu2022softgroup}&51.6&66.1&\textbf{73.6}&\textbf{66.6}\\
    3D-QueryIS&\textbf{55.5}&\textbf{66.7}&68.6&65.6\\
			\midrule
			\multicolumn{5}{c}{Test on 6-fold  cross validation} \\
			\midrule
			SGPN~\cite{wang2018sgpn} &  -  & -  &  38.2  & 31.2   \\
			ASIS~\cite{wang2019asis}&- &-  & 63.6 & 47.5 \\
			3D-BoNet~\cite{yang20193dbonet} &- &- &65.6 &47.6 \\
			PartNet~\cite{yu2019partnet}  &- &- &56.4 &43.4 \\ 
			MPNet~\cite{he2020eccvmemory}  &- &- &{68.4} &{53.7} \\
			
			PointGroup~\cite{jiang2020pointgroup} &- &64.0 &69.6 &69.2 \\
			HAIS~\cite{Chen_HAIS_2021_ICCV} & - &- &73.2 &69.4\\
      MaskGroup~\cite{9859996}&-&65.0&62.7&64.7\\
			SSTNet~\cite{liang2021instance} &54.1 &67.8 &73.5 &\textbf{73.4} \\
			SoftGroup~\cite{vu2022softgroup}&54.4&68.9&\textbf{75.3}&69.8\\
   3D-QueryIS&\textbf{62.2}&\textbf{73.9}&73.2&72.3\\
			\bottomrule
		\end{tabular}
	\end{center}
	 \vspace{-5mm}
	\label{tab:s3dis_ins_results}
\end{table}
\subsection{Ablation Study}
To investigate the effectiveness of different components in our method,
we take our method with a small backbone (i.e., 3D-QueryIS-S) on the validation split of ScanNetV2 as an example to perform comprehensive ablation studies.



\noindent\textbf{The effect of RPG module.}
To investigate the effectiveness of the Representative Point Generation (RPG) module,
we further propose three alternative variants of our 3D-QueryIS (i.e., 3D-QueryIS (w/o RPG), 3D-QueryIS (Random), 3D-QueryIS (FPS)) as shown in Table~\ref{a1}.
Specifically,
for 3D-QueryIS (w/o RPG),
we propose to directly use the generated voxel embeddings of the $t$-th decoder layer from the U-Net decoder (i.e., $\mathbf{O}$) for generating the key and value used in the IQD module.
For 3D-QueryIS (Random),
we randomly sample $J$ voxels from $\mathbf{O}$ to generate the representative points $\mathbf{S}$.
For 3D-QueryIS (FPS),
we utilize the task-agnostic Farthest Point Sampling (FPS)~\cite{qi2017pointnetplusplus} strategy to sample $J$ voxels from $\mathbf{O}$ as $\mathbf{S}$ for ensuring a good coverage of the original set of voxels.
In Table~\ref{a1},
we observe that our 3D-QueryIS  is much better than these three alternative variants,
which demonstrates the effectiveness of our proposed RPG sampling strategy.

\begin{table}[t]
    \centering
    \caption{Results of different methods on the validation set of the ScanNetV2 dataset.
    }
    \label{a1}
    \small
    \begin{tabular}{c|cc}
        \toprule
        Methods  &$\mathrm{AP}$ &$\mathrm{AP}_{50}$ \\
        \hline
        3D-QueryIS & \textbf{41.5}&\textbf{63.3} \\
        	\midrule
        			\multicolumn{3}{c}{Ablation on RPG} \\
        	\midrule
       3D-QueryIS (w/o RPG) & 39.0&59.9 \\
       3D-QueryIS (Random) & 39.3&59.9\\
       3D-QueryIS (FPS) & 40.1&61.1\\
       	\midrule
               			\multicolumn{3}{c}{Ablation on Query Type} \\
                         	\midrule

                   3D-QueryIS (Non-Params) & 39.3&61.5\\

        \bottomrule
    \end{tabular}
\end{table}


\noindent\textbf{The effect of query type.}
In  3D-QueryIS, 
we follow~\cite{cheng2022masked,cheng2021per} to use a set of learnable parameters as the  initial queries $\mathbf{Q}$ in IQD.
To analyze the effect of query type,
following existing 3D object detection methods~\cite{misra2021-3detr,liu2021group},
we report the results using queries sampled from U-Net output  $\mathbf{F}$,
which is referred to as 3D-QueryIS (Non-Params).
Specifically, we use FPS to sample from  $\mathbf{F}$ for obtaining $K$ points as $\mathbf{Q}$.
In Table~\ref{a1},
{we observe that 3D-QueryIS  achieves better results than 3D-QueryIS (Non-Params) with data-dependent queries.
For 3D-QueryIS (Non-Params),
we argue that FPS inevitably introduces massive background points~\cite{he2022pointinst3d},
which causes inferior performance. Although this problem can be partially addressed by increasing $K$, large computation overhead makes it insufficient.
}


\begin{figure*}[!htp]
\begin{center}
\includegraphics[width=\linewidth]{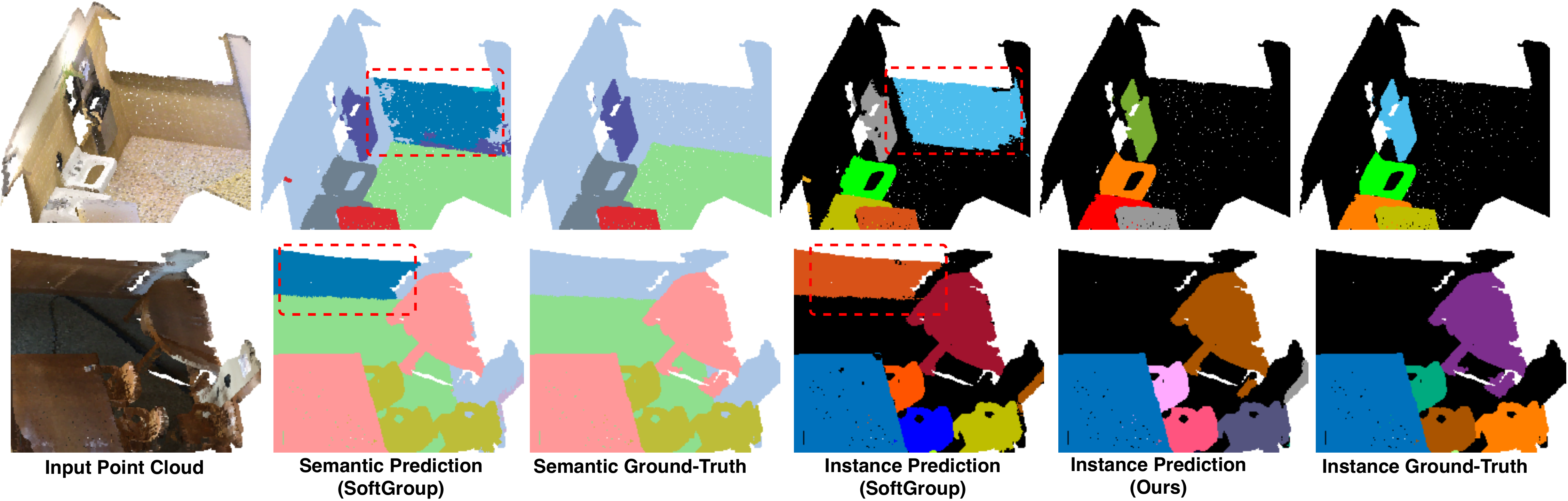}
\caption{
Visualization of different methods.
The instance prediction of SoftGroup depends on the quality of semantic prediction (highlighted by red dashed boxes).
In contrast,
our method directly predicts the instance segmentation without relying on semantic prediction.
}
\label{fig:vis}
\vspace{-4mm}
\end{center}
\end{figure*}
\begin{table}[t]
\setlength\tabcolsep{3pt}
    \centering
    \small
    \caption{Inference time (ms) per scan on the validation set of the ScanNetV2 dataset.}
    \label{Tab:flops}
    \begin{tabular}{c|c|c}
    \toprule
    Methods  & Component time &Total\\ 
    \midrule
     \multirow{3}{*}{PointGroup~\cite{jiang2020pointgroup}} &Initial Feature Extraction (GPU): 55& \multirow{3}{*}{383}\\
     &Clustering (GPU+CPU): 223&\\
     &ScoreNet (GPU): 105&\\
     \hline

     \multirow{3}{*}{HAIS~\cite{Chen_HAIS_2021_ICCV}} &Initial Feature Extraction (GPU): 59 &\multirow{3}{*}{285}\\
                  &Aggregation (GPU+CPU): 136 &\\
     &Refinement (GPU): 90&\\
     \hline
          \multirow{3}{*}{SoftGroup~\cite{vu2022softgroup}} &Initial Feature Extraction (GPU): 58 &\multirow{3}{*}{295}\\
                  &Grouping (GPU+CPU): 141&\\
     &Refinement (GPU): 96&\\
     \hline
    \multirow{3}{*}{3D-QueryIS}& Initial Feature Extraction (GPU): 58&\multirow{3}{*}{\textbf{85}}\\
                      &Instance Query Generation (GPU): 25&\\
     &Instance Mask Prediction (GPU): 2&\\
    \bottomrule
    \end{tabular}
\end{table}

\begin{figure}[t]
\begin{center}
\includegraphics[width=\linewidth]{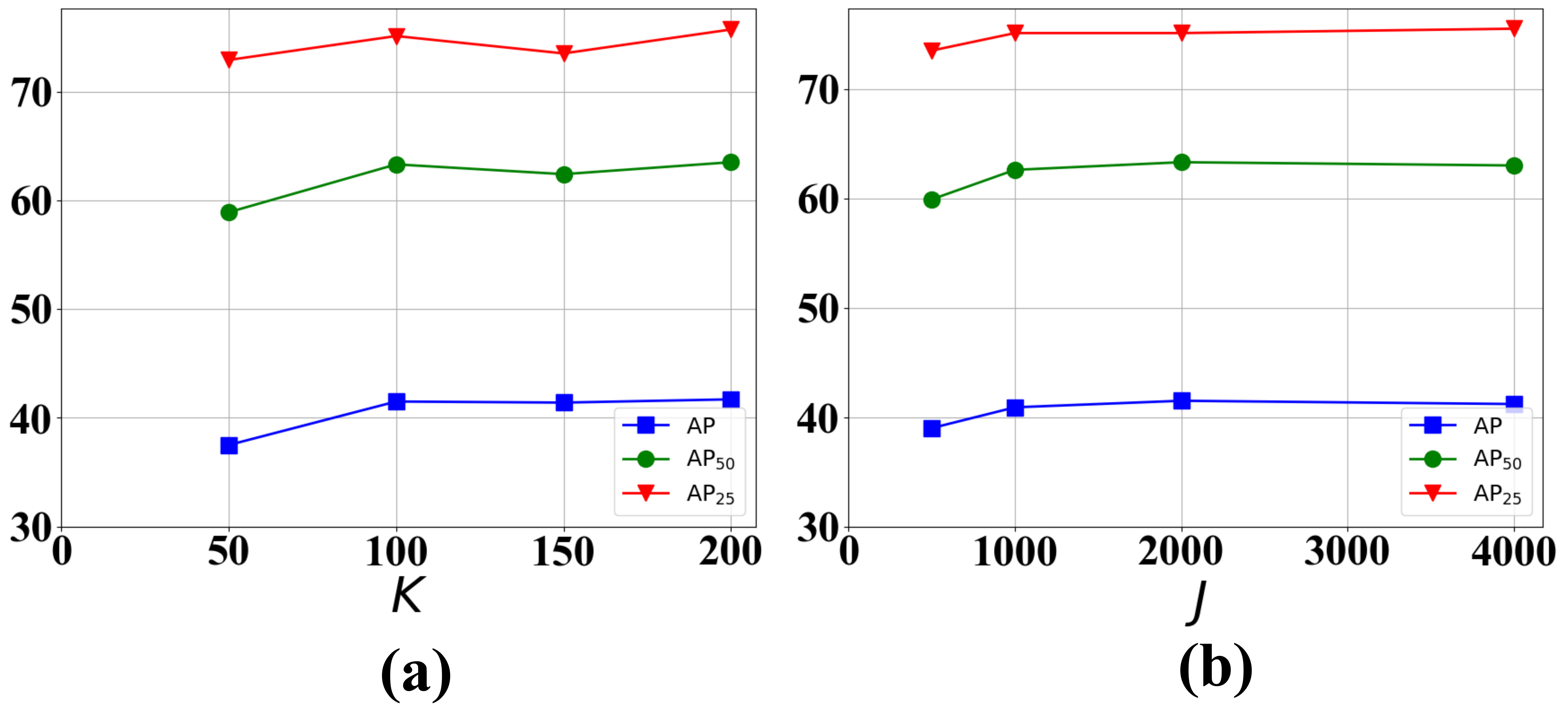}
\caption{
(a). The effect of the number of queries (i.e., $K$).
(b).  The effect of the number of representative points (i.e., $J$).
}
\label{fig:ablation}
\end{center}
\end{figure}

\noindent\textbf{The effect of the number of queries.}
We evaluate our 3D-QueryIS method using  different numbers of queries (i.e., $K$),
and the results on the validation set of ScanNetV2 are shown in Fig.~\ref{fig:ablation}(a).
Specifically,
 when we increase $K$ from 50 to 100,
better performance results are achieved.
When we continue to increase $K$,
the results are relatively stable.
Meanwhile,
the computation cost also increases when $K$ is larger.
Thus,
by default,
we set $K=100$ to achieve the trade-off between performance and efficiency.

\noindent\textbf{The effect of the number of representative points.}
Meanwhile,
we investigate the effectiveness of  our 3D-QueryIS method using  different numbers (i.e., ${J}$) of representative points,
and the results on the validation set of V2 are shown in Fig.~\ref{fig:ablation}(b).
From Fig.~\ref{fig:ablation}(b),
we observe that our proposed 3D-QueryIS achieves better performance when $J$ increases from 500 to 2000.
Meanwhile,
when $J$ continues to increase,
the performance gain vanishes.
Similarly, the computational complexity will also increase as the $J$ increases.
 
\subsection{Further Analysis}
\noindent\textbf{Time Complexity.}
In Table~\ref{Tab:flops},
we compare the inference speed of  3D-QueryIS with existing methods,
where the inference speed is averaged over the  ScanNetV2 validation dataset.
Specifically, the reported runtime is measured on the same A100 GPU,
and we observe 3D-QueryIS is faster than  other methods a lot,
which shows the efficiency of 3D-QueryIS.
Moreover,
regarding the component time, 
existing methods usually  use CPU for clustering or pre-processing (e.g., super-point extraction~\cite{han2020occuseg,liang2021instance}),
while all components of 3D-QueryIS are run on GPU without using CPU.
\noindent\textbf{Visualization on inter-task dependencies.}
In Fig.~\ref{fig:vis},
we provide visualization samples from the ScanNetV2 validation dataset.
Although SoftGoup adopts soft semantic scores to group for mitigating the problems stemming from semantic prediction errors,
the generated instance proposals in SoftGoup still depend on the quality of the semantic branch.
Specifically,
from Fig.~\ref{fig:vis},
for SoftGroup,
we observe that the background (wall) is first considered as foreground instances in semantic prediction, and then instance masks are predicted on the wall in instance prediction,
which means that the semantic prediction
errors are propagated to instance segmentation predictions (highlighted by red dashed boxes).
In contrast, our method removes the inter-task dependencies between the semantic and instance predictions,
and directly generates the instance predictions without relying on semantic predictions.

\begin{figure*}[t]
\begin{center}
\includegraphics[width=\linewidth]{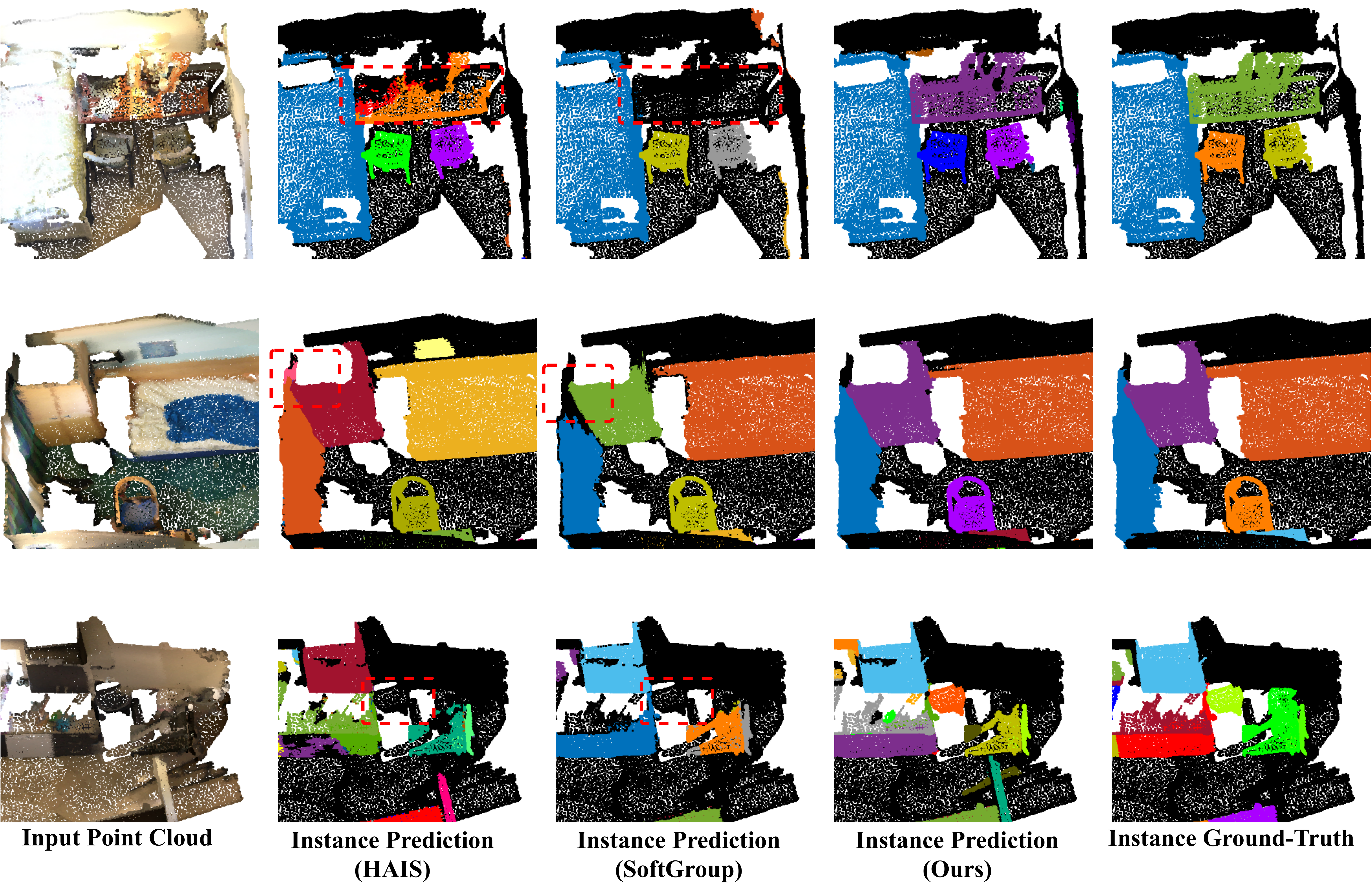}
\caption{
Qualitative results on ScanNet v2 validation set
when compared with the results of HAIS~\cite{Chen_HAIS_2021_ICCV} and SoftGroup~\cite{vu2022softgroup}.
The red boxes highlight the key regions,
and we observe that the prediction results with our proposed 3D-QueryIS shows more accurate instance masks
at these regions.
Zoom in for best view.
}
\label{fig:vis1}
\end{center}
\end{figure*}

\noindent\textbf{Visualization of different methods.}
The visualization of different methods for 3D instance segmentation  is shown in Fig.~\ref{fig:vis1}. 
When compared with the existing methods (e.g., HAIS~\cite{Chen_HAIS_2021_ICCV} and SoftGroup~\cite{vu2022softgroup}), our proposed 3D-QueryIS can correctly segment each instance well and generate finer segmentation results,
which further demonstrates the effectiveness of our proposed 3D-QueryIS method.

\section{Conclusion}
In this paper, we  analyze the drawbacks of existing 3D instance segmentation methods,
and propose a new query-based framework 3D-QueryIS to reduce the inter-task dependencies  for 3D instance segmentation,
where our 3D-QueryIS is detector-free, semantic segmentation-free, and cluster-free.
The 3D-QueryIS consists of initial feature extraction, instance query generation, and instance mask prediction stages. 
Comprehensive experiments on multiple benchmark datasets demonstrate the effectiveness and efficiency of our 3D-QueryIS method.
In the future,
Moreover,
we hope our method can motivate other researchers to further investigate the query-based framework for better performance and efficiency on 3D instance segmentation.

{\small
\bibliographystyle{ieee_fullname}
\bibliography{egbib}
}

\end{document}